\documentclass[conference]{IEEEtran}
\IEEEoverridecommandlockouts
\usepackage{cite}
\usepackage{amsmath,amssymb,amsfonts}
\usepackage{algorithmic}
\usepackage{graphicx}
\usepackage{textcomp}
\usepackage{xcolor}
\usepackage{newtxmath}
\def\BibTeX{{\rm B\kern-.05em{\sc i\kern-.025em b}\kern-.08em
    T\kern-.1667em\lower.7ex\hbox{E}\kern-.125emX}}
    
\usepackage{natbib} 
\usepackage{mathtools} 
\usepackage{booktabs} 
\usepackage{tikz} 

\usepackage[]{amsmath}    
\usepackage[]{graphicx}    
\usepackage{subfig}
\usepackage{hyperref}
\usepackage{amssymb}
\usepackage{dsfont}

\newcommand{\ignore}[1]{}  

\begin{document}

\title{Comparing the quality of neural network uncertainty estimates for classification problems \thanks{\copyright2023 IEEE. Personal use of this material is permitted.  Permission from IEEE must be obtained for all other uses, in any current or future media, including reprinting/republishing this material for advertising or promotional purposes, creating new collective works, for resale or redistribution to servers or lists, or reuse of any copyrighted component of this work in other works. This is the accepted version of the article published available at \url{https://doi.org/10.1109/ICMLA55696.2022.00039}}}

\author{\IEEEauthorblockN{Daniel Ries}
\IEEEauthorblockA{\textit{Statistics and Data Analytics} \\
\textit{Sandia National Laboratories}\\
Albuquerque, USA \\
dries@sandia.gov}
\and
\IEEEauthorblockN{Joshua Michalenko}
\IEEEauthorblockA{\textit{Proliferation Signature and Data Exploitation} \\
\textit{Sandia National Laboratories}\\
Albuquerque, USA \\
jjmich@sandia.gov}
\and
\IEEEauthorblockN{Tyler Ganter}
\IEEEauthorblockA{\textit{Applied Machine Intelligence} \\
\textit{Sandia National Laboratories}\\
Albuquerque, USA \\
tganter@sandia.gov}
\and
\IEEEauthorblockN{Rashad Imad-Fayez Baiyasi}
\IEEEauthorblockA{\textit{Mission Algorithms R\&S} \\
\textit{Sandia National Laboratories}\\
Albuquerque, USA \\
ribaiya@sandia.gov}
\and
\IEEEauthorblockN{Jason Adams}
\IEEEauthorblockA{\textit{Statistical Sciences} \\
\textit{Sandia National Laboratories}\\
Albuquerque, USA \\
jradams@sandia.gov}
}


\thispagestyle{plain}
\pagestyle{plain}

\maketitle




\begin{abstract}
Traditional deep learning (DL) models are powerful classifiers, but many approaches do not provide uncertainties for their estimates. Uncertainty quantification (UQ) methods for DL models have received increased attention in the literature due to their usefulness in decision making, particularly for high-consequence decisions. However, there has been little research done on how to evaluate the quality of such methods. We use statistical methods of frequentist interval coverage and interval width to evaluate the quality of credible intervals, and expected calibration error to evaluate classification predicted confidence. These metrics are evaluated on Bayesian neural networks (BNN) fit using Markov Chain Monte Carlo (MCMC) and variational inference (VI), bootstrapped neural networks (NN), Deep Ensembles (DE), and Monte Carlo (MC) dropout. We apply these different UQ for DL methods to a hyperspectral image target detection problem and show the inconsistency of the different methods' results and the necessity of a UQ quality metric. To reconcile these differences and choose a UQ method that appropriately quantifies the uncertainty, we create a simulated data set with fully parameterized probability distribution for a two-class classification problem. The gold standard MCMC performs the best overall, and the bootstrapped NN is a close second, requiring the same computational expense as DE. 
Through this comparison, we demonstrate that, for a given data set, different models can produce uncertainty estimates of markedly different quality. This in turn points to a great need for principled assessment methods of UQ quality in DL applications.

\end{abstract} 


\section{Introduction}

Traditional deep learning (DL) models are powerful predictors in both regression and classification problems (\cite{lecun2015}), but many do not provide uncertainties for their predictions or estimates.  The usefulness of uncertainty quantification (UQ) in DL models is being recognized, especially for applications that are high-consequence, including nuclear stockpile stewardship and safety (\cite{trucano2004,stracuzzi2018}), nuclear energy (\cite{stevens2016}), national security problems (\cite{ries2022,gray2022}), and medical diagnoses (\cite{begoli2019,kompa2021}). For example, \cite{kompa2021} explains the benefit of using UQ in medical decision making, including models that can report ``I don't know'' to ensure human experts will further evaluate results.

\subsection{High Consequence Application}

Hyperspectral images (HSI) contain information across hundreds of spectral bands over a surface. These spectral bands provide crucial information about what is in the scene, giving significantly more information than the human eye can detect. A common application of HSI is target detection, where an observer is trying to determine if an object of interest is in the image (\cite{nasrabadi2013,poojary2015,anderson2019}). Of particular interest for national security problems is finding rare or hidden targets. Past work (\cite{anderson2019,gray2022}) has shown the ability to detect targets at the sub-pixel level. However, the high consequence nature of target detection  applications have an extremely high cost for false positives where the need for trustworthy algorithms is paramount. Uncertainty quantification of model predictions is becoming a necessity in high consequence problems (\cite{trucano2004,begoli2019}) to help alleviate this problem. Traditional DL methods only provide a best estimate, and do not provide an estimate of the model's confidence in its predictions. \cite{ries2022} applied Bayesian neural networks (BNN) to an HSI target detection problem and proposed High Confidence sets (HCS) as a way to operationalize UQ output. There are many ways (other than BNNs) to quantify model uncertainty, and the decision maker must determine which UQ approach is most representative of the true uncertainty. 

Comparing different UQ methods on this application, we clearly demonstrate that, for a given data set, different models can produce uncertainty estimates of markedly different quality. This in turn points to a great need for principled methods to assess UQ quality in DL applications.

\subsection{UQ Methods for DL Models}
\label{sec:uq4dl}

Bayesian neural networks were first popularized by David MacKay (\cite{mackay1992,mackay1995}) and his student Radford Neal (\cite{neal1996}). Neal's dissertation introduced Hamiltonian Monte Carlo (HMC) as a way to sample the posterior distribution of a BNN, providing a practical way of training using Markov Chain Monte Carlo (MCMC). To this day, HMC is considered the gold standard for BNN training due to its theoretical backing and lack of approximations. Interested readers should consult \cite{gelman2013} for more details and references about MCMC and HMC.

Variational inference is the most popular method of Bayesian inference for neural networks (NN) (\cite{graves2011}). \cite{blei2017} gives an extensive review of VI methods. Although VI is computationally much cheaper than MCMC, a common criticism of standard implementations of VI is the mean-field assumption (assuming posterior independence of all parameters). Put simply, VI is an approximation to the posterior distribution using optimization that improves as the sample size  increases, compared to MCMC which is an approximation to the posterior distribution using sampling that improves as the number of Monte Carlo (MC) samples increases. Therefore, VI is constrained by data, and MCMC is constrained by computation time.  

The bootstrap is a simulation-based method that treats the training data as the population and samples new data sets with replacement from the original training set. Uncertainty is measured by creating a large number of these new data sets and then using the distribution of estimates or predictions to quantify uncertainty (\cite{gray2022}). Deep Ensembles (DE)  (\cite{laka2017}) follow a similar idea to the bootstrap except no resampling is done; the only difference for each model in the ensemble is the set of starting values for the model optimizer. Monte Carlo  Dropout, proposed by \cite{gal2016}, is an extension of dropout regularization (\cite{srivastava2014}) that understands dropout as a sampling method that approximates a deep Gaussian process (GP). Unlike traditional dropout regularization, which is only applied during training, MC Dropout includes dropout during inference. In this way, an ensemble of predictions can be obtained from a single trained NN, allowing for uncertainty to be estimated. Comprehensive reviews of UQ methods in DL can be found in \cite{kabir2018} and \cite{abdar2021}.

\subsection{Review of assessing quality of UQ in DL}

Unlike evaluating a DL model's predictive performance using metrics like mean squared error (MSE) or accuracy, a commonly accepted UQ quality metric does not exist, but some previous work has sought to address this problem.  \cite{kabir2018} reviews the ideas of frequentist coverage and interval width as tools for UQ evaluation and cites several examples.  \cite{yao2019} evaluates the predictive uncertainty for several BNN training methods and ensembles. The authors found ensembles do not provide the UQ that users believe it provides, and emphasize calibration metrics are not good indicators of posterior approximation. The authors concluded a new metric for assessing predictive uncertainty is needed.  \cite{ovadia2019} gives a large-scale benchmark of current UQ for DL methods using metrics such as negative log likelihood, Brier score, and expected calibration error (ECE). The authors find many methods have trouble with out of distribution (OOD) situations or with dataset shift.  \cite{stahl2020} evaluated several UQ for DL methods, including BNN and DE, and found they captured the uncertainty differently and correlations between the methods' quantifications were low.  \cite{kompa2020} checked empirical frequentist coverage and interval widths for several DL methods. The authors found MC  dropout and ensembling to have low interval coverages and high variability in results on a regression example. In comparison, BNN and GP provided the expected coverages and low variability in the results. For classification, all methods gave adequate coverages for independent and identically distributed (i.i.d.) data, but methods generally performed poorly in terms of coverage when dataset shift was added.   \cite{naeini2015} developed the Expected Calibration Error (ECE) metric for classification models which assesses the agreement of predicted confidences and model accuracy. \cite{nado2021} baselines UQ quality using ECE and creates a user-friendly framework for assessing performance across multiple UQ methods and model architectures, but the authors do not address epistemic uncertainty. 

A desired metric to compare and assess uncertainty estimates should consider both aleatoric and epistemic uncertainties. In brief, aleatoric uncertainty is the variability due to randomness or noise in the process or measurement. This type of uncertainty is always present and can only be reduced by an improvement in the process of measurement, not by increasing the sample size. Epistemic uncertainty is the uncertainty resulting from imperfect knowledge of the model. Examples of this include uncertainty during model selection and parameter uncertainty during training. Increasing sample sizes will help reduce epistemic uncertainty by either further understanding the mechanism and creating better model architectures, estimating model parameters more precisely, or both. A comprehensive introduction to the two types of uncertainties in the context of machine learning is given by \cite{hullermeier2021}. 




This paper is organized as follows: Section 2 introduces the motivating application and presents results which necessitate further exploration. 
Section 3 introduces interval coverage, interval width and ECE, the UQ metrics used in this paper to assess UQ quality. Section 4 applies the metrics in Section 3 on DL models to a simulated classification data set. Finally, Sections 5 and 6 discuss the results and provide conclusions, respectively.

\section{Motivating Application}
\label{sec:application}

Our interest in the quality of the UQ given by a model stems from a target detection problem in a high-consequence decision space described in this section. 

\subsection{Data}

\begin{figure}[h]
\centering
\includegraphics[width=0.3\textwidth]{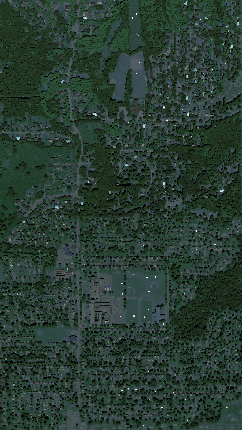}
\caption{Pseudo color render of Megascene MLS-1200 at R=670 nm, G=540 nm, B=480 nm. Image reproduced from \protect\cite{anderson2019}}
\label{scene}
\end{figure}

The synthetic dataset Megascene (\cite{ientilucci2003}), is a high fidelity HSI simulation scene representing a suburban area of Rochester, NY. The scene contains both natural and man-made objects. Figure \ref{scene} shows a pseudo color rendering of the entire scene, designated as MLS-1200; roads, houses, trees, and even a track can be seen in the image. The simulator uses an AVIRIS-like sensor measuring 211 spectral bands ranging from 0.4 to 2.5 $\mu m$. The images were created such that the scene is being observed at an elevation of 4 km, giving a pixel size of 1 m$^2$. Therefore at each pixel, we have the complete spectrum from 0.4 to 2.5 $\mu m$, and we know exactly the contents of that pixel which make up the spectrum. Details about the radiance to reluctance conversion, in addition to other specifics, can be found in \cite{anderson2019}.

\begin{figure}[h]
\includegraphics[width=0.5\textwidth]{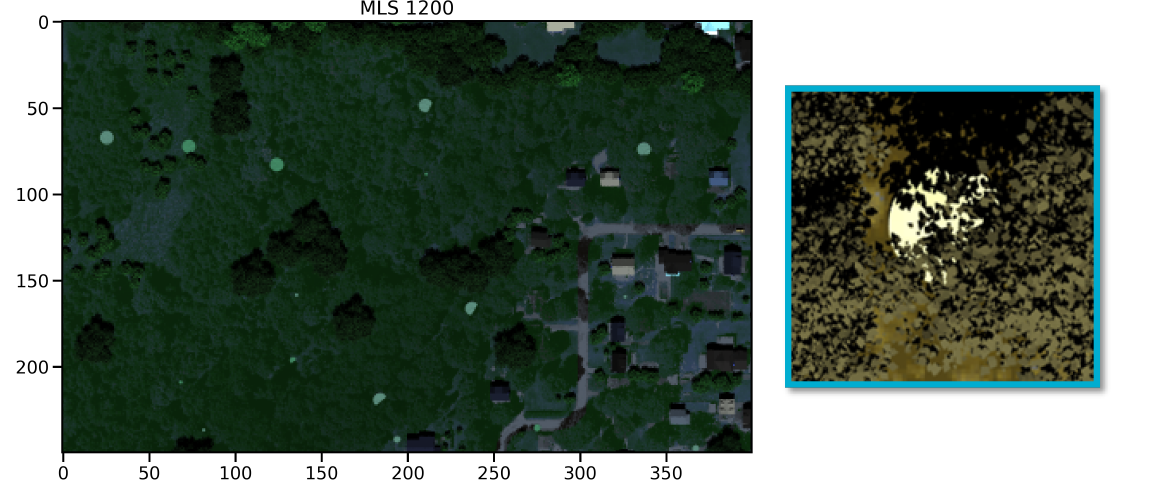}
\caption{Left: Subset of Megascene showing inserted target green discs. Right: Example of a green disc partially hidden by foliage. Image reproduced from \protect\cite{anderson2019}. }
\label{targets}
\end{figure}

We are interested in detecting small targets hidden within a scene. We manually inserted green discs (with a known spectrum) randomly through the scene to represent targets to detect. In total, the scene contains 125 discs ranging in size from 0.1 to 4m radii. Given the pixel size of 1 $m^2$, some targets fill multiple pixels while others fill just a fraction of a pixel. To make the targets more realistic, some of the discs were partially hidden beneath foliage. Figure \ref{targets} (Figure 6 in \cite{anderson2019}), shows a subset of Megascene with several different sized green target discs. The image on the right shows an example of a disc partially hidden by foliage. 

\subsection{Training}

Several methods were described in Section \ref{sec:uq4dl} which provide the necessary UQ for high consequence applications, any of which would be valid for this application. The architecture of the neural networks is 2 hidden layers with 10 nodes each. The left half of MLS-1200 was used for training, and the right half was used for testing. 

Let $\mathcal{D} = \{({\bf x}_i,y_i)\}_{i=1}^n$, be the training data set where ${\bf y} = (y_1,...y_n)'$ and ${\bf X} = ({\bf x}_1,...,{\bf x}_n)'$. Let $y_i \in \{0,1\}$, denoting non-target or target and ${\bf x}_i \in \mathbb{R}^{p}$ be a $p$-dimensional vector of features corresponding to response $y_i$. Let $\boldsymbol{\theta}$ denote all the weights and biases of the neural network. The neural network $\pi: \mathbb{R}^{p} \rightarrow (0,1)$ estimates the probability that pixel $i$ contains target as, $\pi_i = P(y_i=1|{\bf x}_i,\boldsymbol{\theta})$.


\subsection{Quantifying Uncertainty}

Uncertainty on the neural network is measured on its estimates $\hat{\pi}_i$, which use the trained  models' weights and biases $\hat{\theta}$. The uncertainty of $\hat{\pi}_i$ is obtained in the form of $(1-\alpha)\%$ credible intervals (CIs), denoted by $\mathcal{B}_{\pi_i}(\alpha)$. 

In order to reduce analyst burden through automation, we want to know where the model believes, with high-confidence, whether or not a pixel contains a target. The High Confidence Sets (HCS) proposed in \cite{ries2022} provide a means to operationalize such a process. Formally, the HCS $\Omega$ is the set of pixels such that:

\begin{align}
        \Omega &= \{i: (\mathcal{B}_{\pi_i}(\alpha)_{LB} > 1-\delta  \cup  \mathcal{B}_{\pi_i}(\alpha)_{UB} < \delta)\}
    \label{def:hc}
\end{align}

where $\mathcal{B}_{\pi_i}(\alpha)_{LB}$ and $\mathcal{B}_{\pi_i}(\alpha)_{UB}$ are the lower and upper bounds of a $(1-\alpha)$\% CI for $\pi_i$, respectively; $\delta$ is a probability threshold which defines an estimated probability as close to zero. Both $\alpha$ and $\delta$ are user chosen and should reflect the users' risk preferences. We choose $\alpha=\delta=0.2$.

 Table \ref{tab:megasceneresults} shows the proportion of pixels from the test set which were included in the respective HCS. There are clear differences in the results, begging the question: which UQ method should the decision maker rely on? In the test scene with over 1.5 million pixels, the 10\% difference between BNN-MCMC and DE corresponds to a difference in HCS size of 150,000 pixels. This difference can have a large effect on analysts, but the UQ method with the largest HCS should not automatically be relied on since it could be overconfident. An UQ quality assessment is needed.

\begin{table}[t]
\begin{tabular}{l|c}
\hline
Method & Proportion of Pixels in HC Set \\
\hline
BNN-MCMC & 0.81 \\
BNN-VI & 0.27  \\
DE & 0.71 \\
Bootstrap & 0.78 \\
MC Dropout & 0.74 \\
\hline
\end{tabular}
\caption{Proportion of test set pixels in HCS for Megascene for each model.}
\label{tab:megasceneresults}
\end{table}

\section{Uncertainty Quantification Quality Metrics}
\label{sec:quality_metrics}

This section introduces UQ metrics that can be used to evaluate UQ model performance and help answer the question posed at the end of the previous section. Some of these metrics require knowing the complete probabilistic data generating mechanism, which in real data problems is not generally known. Therefore the simulation study in Section \ref{sec:simstudy} is needed to evaluate the different UQ methods used in Section \ref{sec:application}.

\subsection{Frequentist Interval Coverage}

Credible intervals  contain a set of plausible class probability estimates, where plausible is defined by the \emph{nominal} rate of the interval itself, typically denoted as $(1-\alpha$)\%. A $(1-\alpha)$\% CI for an estimate should contain the true population parameter about $(1-\alpha)$\% of the time if the experiment was redone. Frequentist coverage (coverage, from here on) is the \emph{actual} rate at which the population parameter is contained in the interval, averaged over all observations. 

\begin{align}
 \text{CI Coverage} &= \frac{1}{n} \sum_{i=1}^n \mathds{1}\big(\pi_i \in \mathcal{B}_{\pi_i}\text{}(\alpha)\big)
 \label{eq:ci_coverage}
\end{align}

This empirical value should be as close as possible to the nominal rate of $(1-\alpha)$\%. Going under or over this value is an indication of poor UQ quality, e.g. a 90\% CI with 70\% coverage indicates the interval is overly optimistic and not accounting for enough uncertainty. Conversely a 90\% interval with 99\% coverage is overly conservative. Note that Equation \eqref{eq:ci_coverage} requires knowing the \emph{true} value of the model parameter.

\subsection{Interval Width}

Intervals contain values that are plausible estimates for a quantity of interest, therefore it would make sense that there is less variability in the data generating mechanism if the interval is smaller. However, it is not quite this simple.   The highest UQ quality is given to models that minimize interval width \emph{and} match coverage with nominal rate. The width of intervals is given in Equation \eqref{eq:width} by

\begin{equation}
 \text{Interval Width} = \frac{1}{n} \sum_{i=1}^n (\mathcal{B}_{\pi_i}(\alpha)_{UB}  - \mathcal{B}_{\pi_i}(\alpha)_{LB}).
 \label{eq:width}
\end{equation}


\subsection{Expected Calibration Error}

\cite{naeini2015} proposed ECE as a metric to check whether a machine learning classifier's confidence scores are calibrated to true probabilities of correctness. 
Here we use the broader term \emph{predicted confidence} defined as $\hat{\pi}_i \equiv  \pi({\bf x}_i,\hat{\theta}) \in [0,1]$, or estimated class probabilities. However, we make no claim that all models are expected to estimate the true probability. 
For classification BNNs, the uncertainty of interest is on the estimated class probabilities (predicted confidences).

Consider a binary decision rule, $\tau(\cdot)$, that generates predictions $\tau(\hat{\pi}_i) = \hat{y}_i \in \{0,1\}$. Provided a set of true and predicted responses, the accuracy is computed as:

\begin{equation}
 acc({\bf y}, {\bf \hat{y}}) = \frac{1}{n} \sum_{i=1}^n \mathds{1}(\hat{y}_i = y_i).
 \label{eq:acc}
\end{equation}

The average confidence of the set is

\begin{equation}
 conf(\hat{\boldsymbol{\pi}}) = \frac{1}{n} \sum_{i=1}^n \hat{\pi}_i.
 \label{eq:conf}
\end{equation}

ECE discretizes the interval [0, 1] under equally spaced bins and assigns each predicted confidence to the bin that encompasses it. The calibration error of a bin is the difference between the accuracy and average confidence of the samples assigned to that bin. In other words, calibration error treats predicted confidences as estimated probabilities and measures the disagreement between the estimated and true probability of correctness. ECE is a weighted average across all bins:

\begin{equation}
ECE({\bf y}, \hat{\boldsymbol{\pi}}) = \sum_{b=1}^B \frac{n_b}{n} \Big| acc\big({\bf y}_b, \tau(\hat{\boldsymbol{\pi}}_b)\big) - conf(\hat{\boldsymbol{\pi}}_b) \Big|.
\label{eq:ece}
\end{equation}

where $B$ is the number of bins, $({\bf y}_b, \hat{\boldsymbol{\pi}}_b)$ is the subset of $({\bf y}, \hat{\boldsymbol{\pi}})$ in the $b^{th}$ bin, and $n_b$ is the number of predictions in bin $b$, i.e. the rank of $\hat{\boldsymbol{\pi}}_b$.

Calibration informs us of the probability of correctness, regardless of cause. However, as model accuracy approaches the limit of irreducible error, calibrated confidences will approach the true probability of the most probable class. As such, calibration error can effectively assess the quality of aleatoric uncertainty estimation. On the other hand, interval coverage and width provide an assessment of epistemic uncertainty in classification problems because the credible intervals should converge to point predictions as estimates of the class probabilities approach the true probabilities.

\section{Simulation Study}
\label{sec:simstudy}

\begin{figure}[h]
\includegraphics[width=0.4\textwidth]{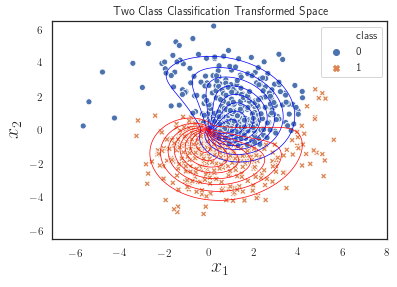}
\caption{TCC transformed space with 10\% contours for $P(Y=y|x_1,x_2)$.}
\label{fig:TC_transformed_space}
\end{figure}

In this section we evaluate UQ metrics of Section \ref{sec:quality_metrics} on a simulated two-class classification (TCC) dataset  to compare  different UQ in DL methods, including BNN trained via MCMC, BNN trained via VI, bootstrapped NN, DE, and MC dropout. For comparison against a non-DL model, we also train a GP  with MCMC.  The TCC dataset is a fully parameterized generative model with a joint probability that allows direct evaluation of CI coverage. A full probability distribution is needed in classification problems to check CI coverage. The underlying model is a 2-D Gaussian Mixture Model (GMM) with two equally proportioned clusters that undergo a series of transformations and scalings. The result is a data model that can easily generate a large variety of data classification scenarios that arise in quantifying UQ. Figure \ref{fig:TC_transformed_space} shows one simulated TCC data set and densities. In all, 100 data sets from the same TCC simulator are generated with each of the UQ methods fit to each of the 100 simulated data sets. Coverage and widths of 90\% credible intervals are computed for each data set for each method, then averaged over the 100 simulations. For the ensemble methods (DE, bootstrap, MC dropout), 100 ensembles were used. The architecture for the DL models was a two layer fully connected NN with 10 nodes per layer.

Table \ref{tab:simresults} shows mean coverage, width, and ECE for each method with its MC standard error in parentheses. Bolded terms show the best metric in each column. Overall, BNN-MCMC does the best since it is the only method to correctly capture the nominal coverage of 0.9. Bootstrap is a close second since it slightly undercovers nominal and has wider intervals than BNN-MCMC. Interestingly, while DE has a coverage rate much less than nominal, its ECE is comparable with BNN-MCMC and bootstrap. This could lead to an erroneous conclusion that DE's UQ is high quality, when in fact it is only \emph{calibrated}, meaning its aleatoric uncertainty is accurate, but based on coverage, its epistemic uncertainty is not. MC Dropout appears to help the ensemble, but it still doesn't achieve nominal coverage. 

\begin{table}[t]
\begin{tabular}{l|ccc}
\hline
Method & Coverage & Width & ECE \\
\hline
BNN-MCMC & {\bf 0.91 (0.04)} & ${\bf 0.22 (0.01)}^*$ & {\bf 0.04 (0.01)} \\
BNN-VI & 0.59 (0.17) & 0.38 (0.07) & 0.08 (0.02) \\
DE & 0.48 (0.09) & 0.09 (0.01) & {\bf 0.04 (0.01)} \\
Bootstrap & 0.84 (0.06) & 0.25 (0.02) & {\bf 0.04 (0.01)} \\
MC Dropout & 0.67 (0.08) & 0.15 (0.02) & {\bf 0.04 (0.01)} \\
GP & 0.98 (0.02) & 0.36 (0.02) & 0.05 (0.01) \\
\hline
\end{tabular}
\caption{TCC Simulation results. Bolded values indicate best metric in each column. The asterisk indicates the best interval width, given the nominal coverage was met (nominal rate = 0.9).}
\label{tab:simresults}
\end{table}

\begin{figure*}[h]
\includegraphics[width=0.3\textwidth]{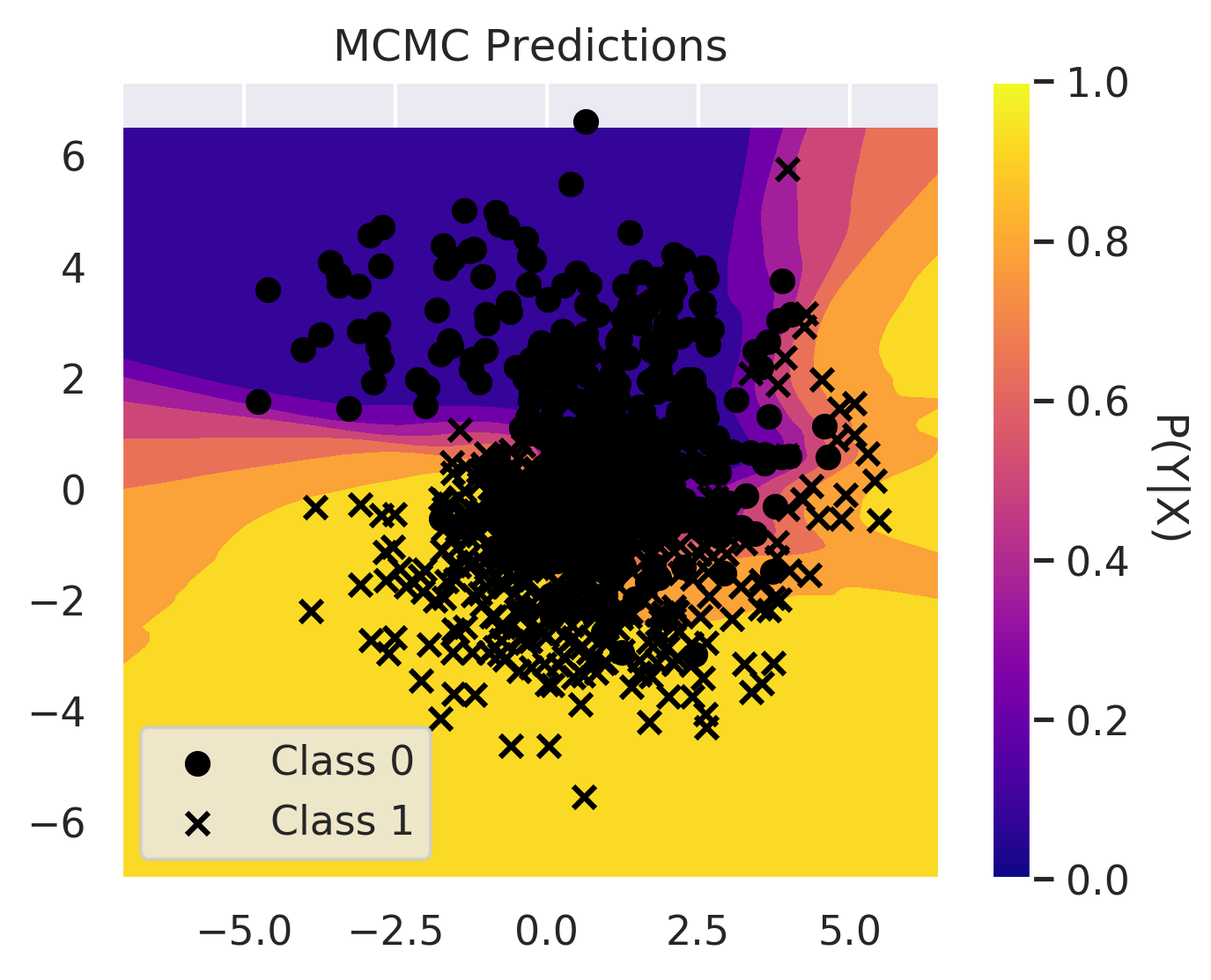} \hfill
\includegraphics[width=0.3\textwidth]{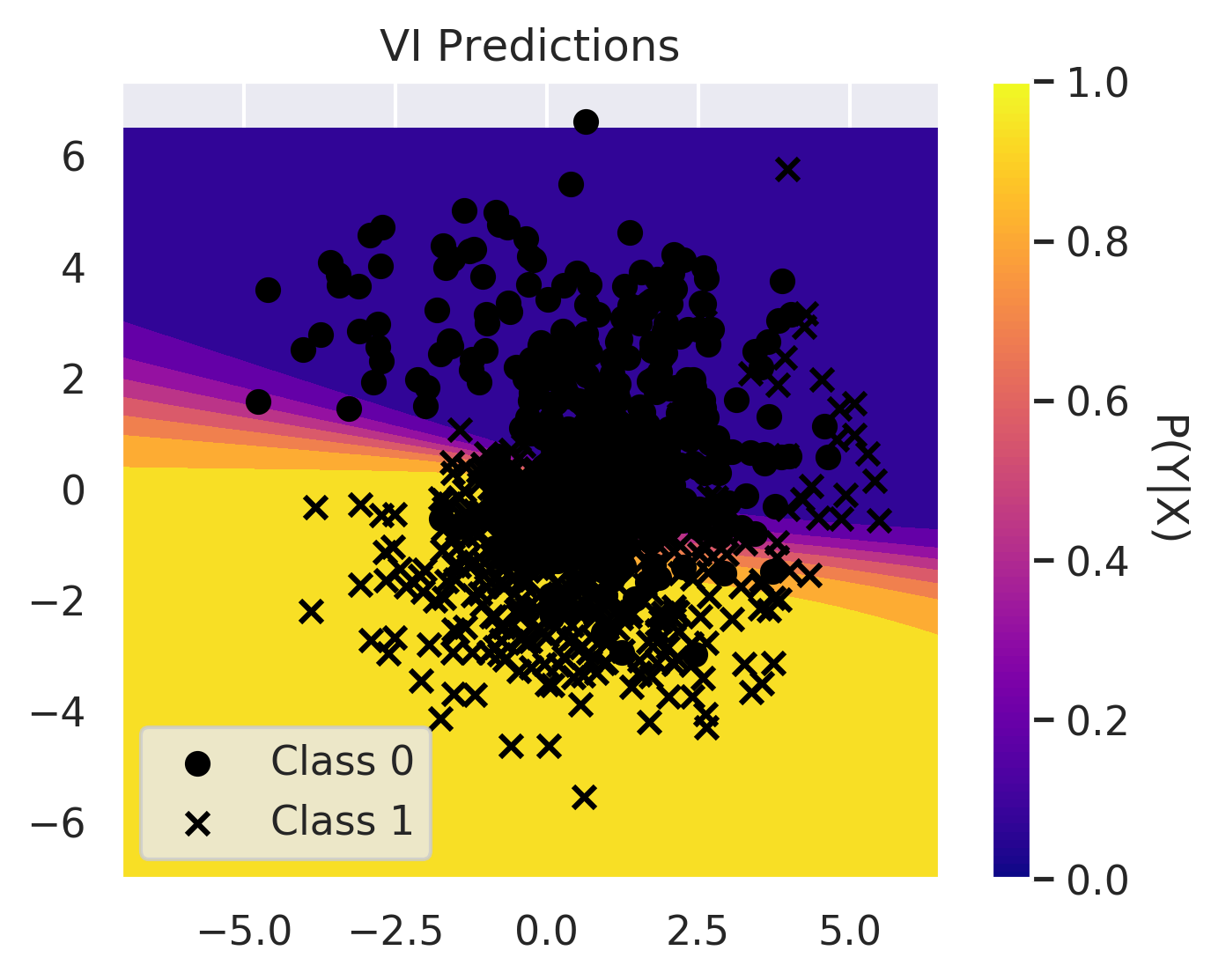} \hfill 
\includegraphics[width=0.3\textwidth]{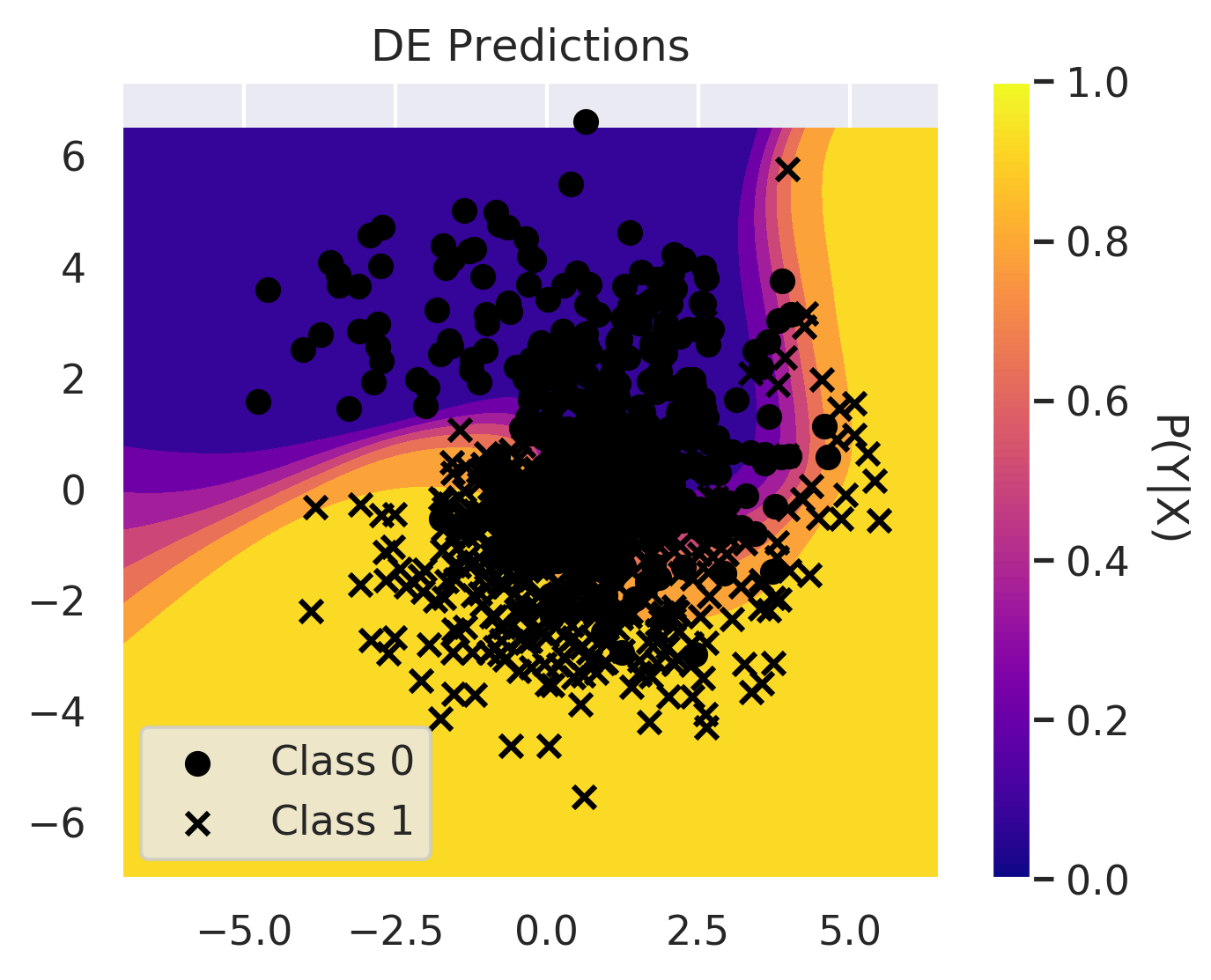} \hfill 
\includegraphics[width=0.3\textwidth]{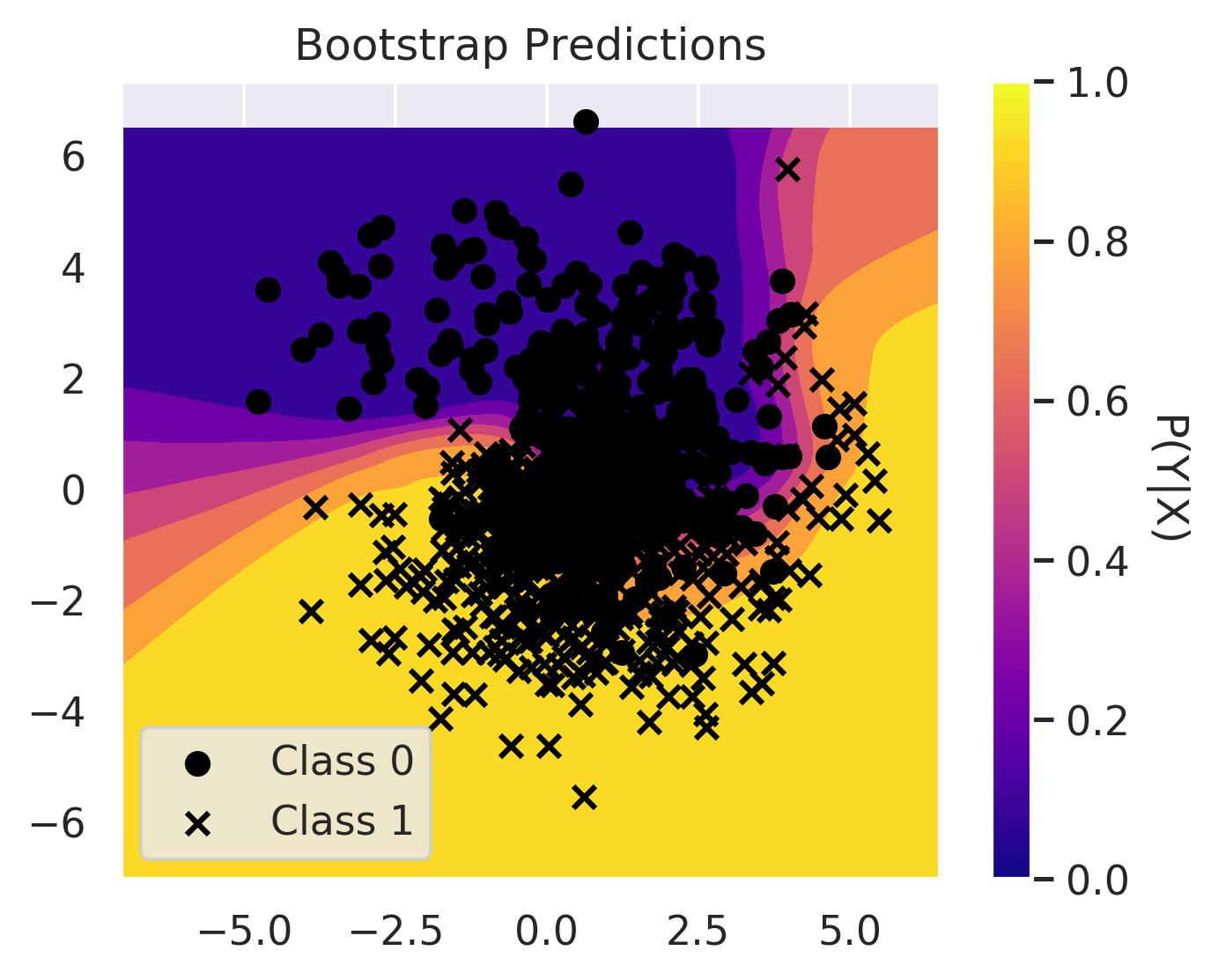} \hfill 
\includegraphics[width=0.3\textwidth]{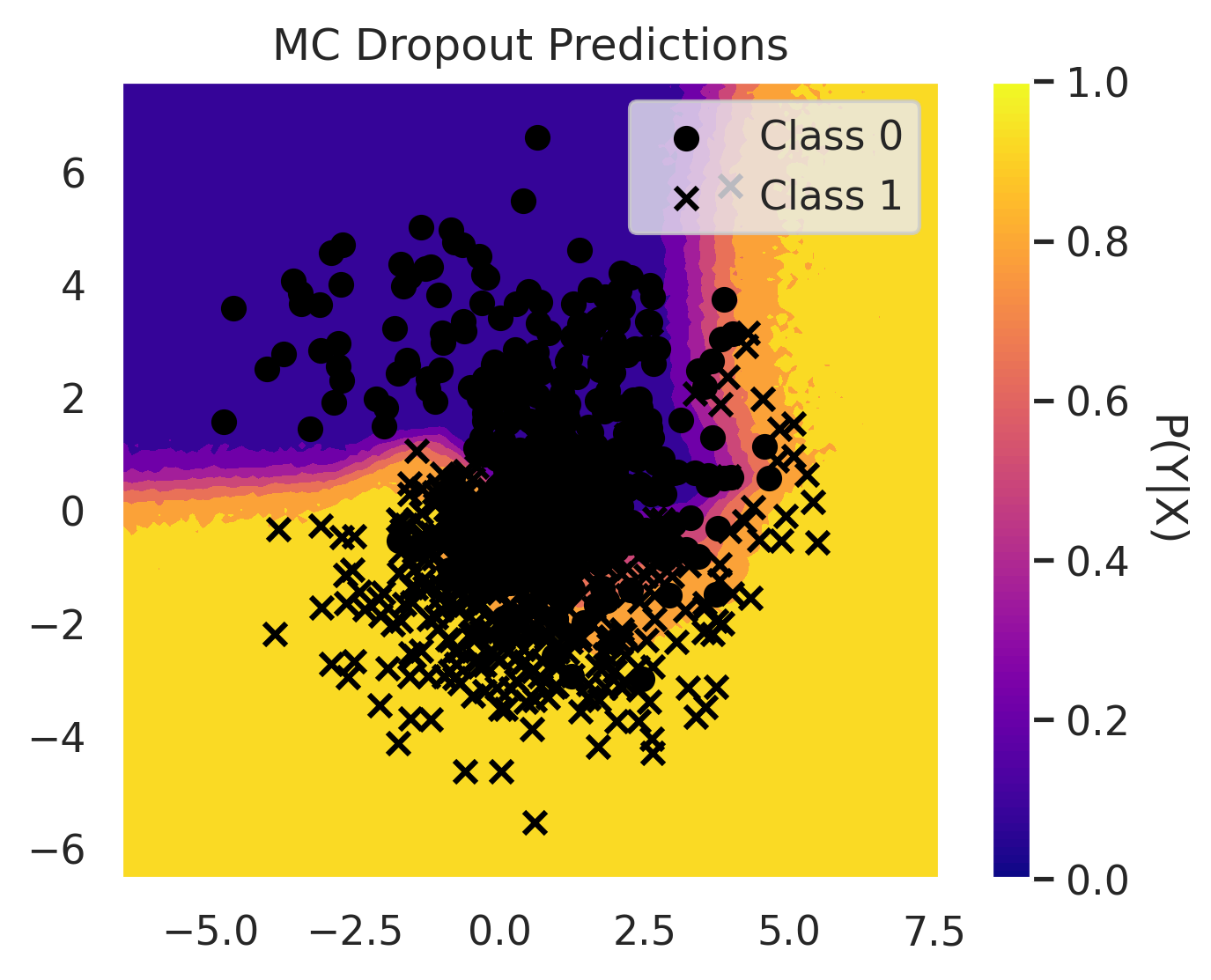} \hfill 
\includegraphics[width=0.3\textwidth]{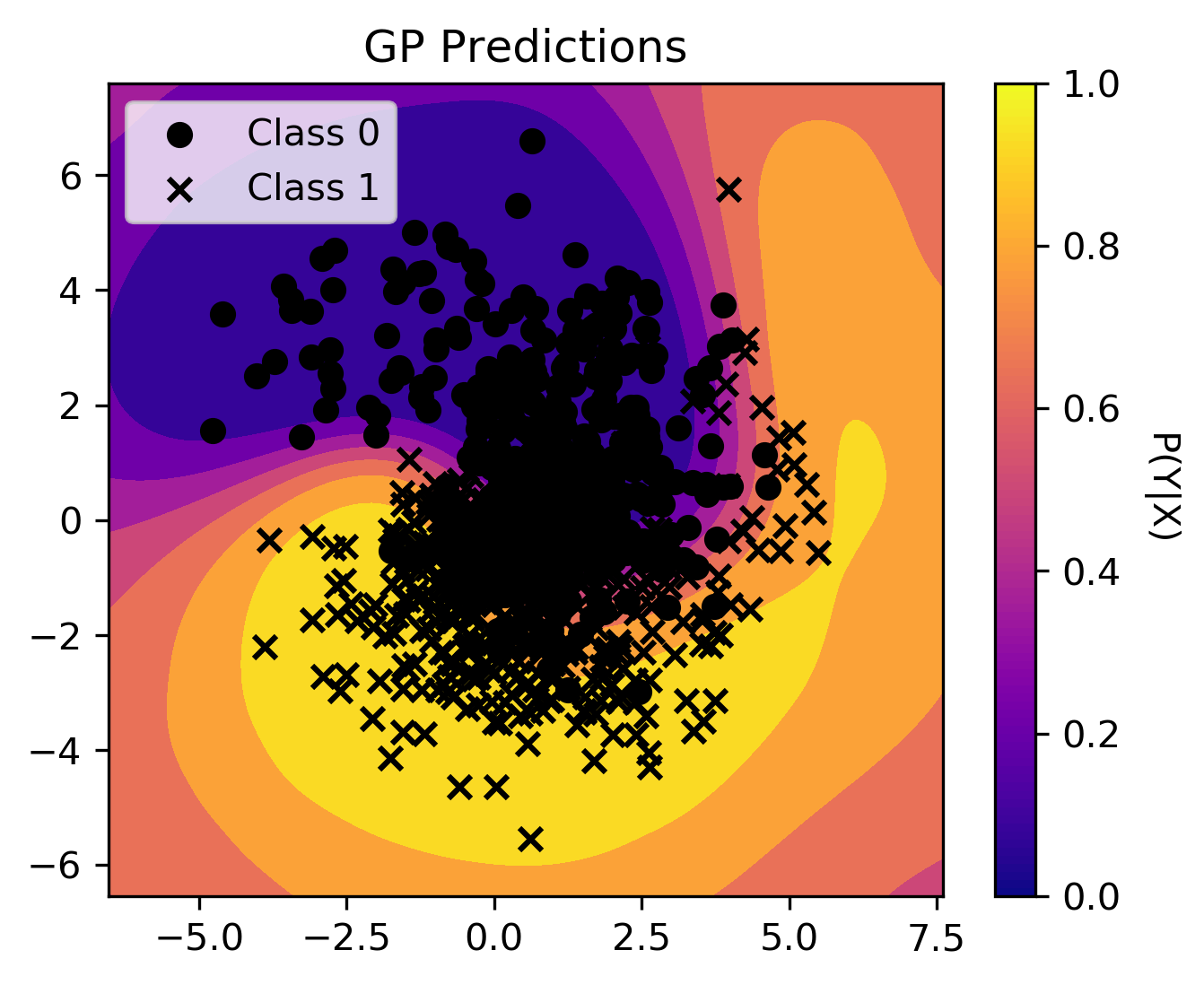} \hfill 
\caption{Prediction surfaces for each model on one TCC simulation. Training data is overlaid.}
\label{fig:TC_mcmc_pred}
\end{figure*}

\begin{figure*}[h]
\includegraphics[width=0.3\textwidth]{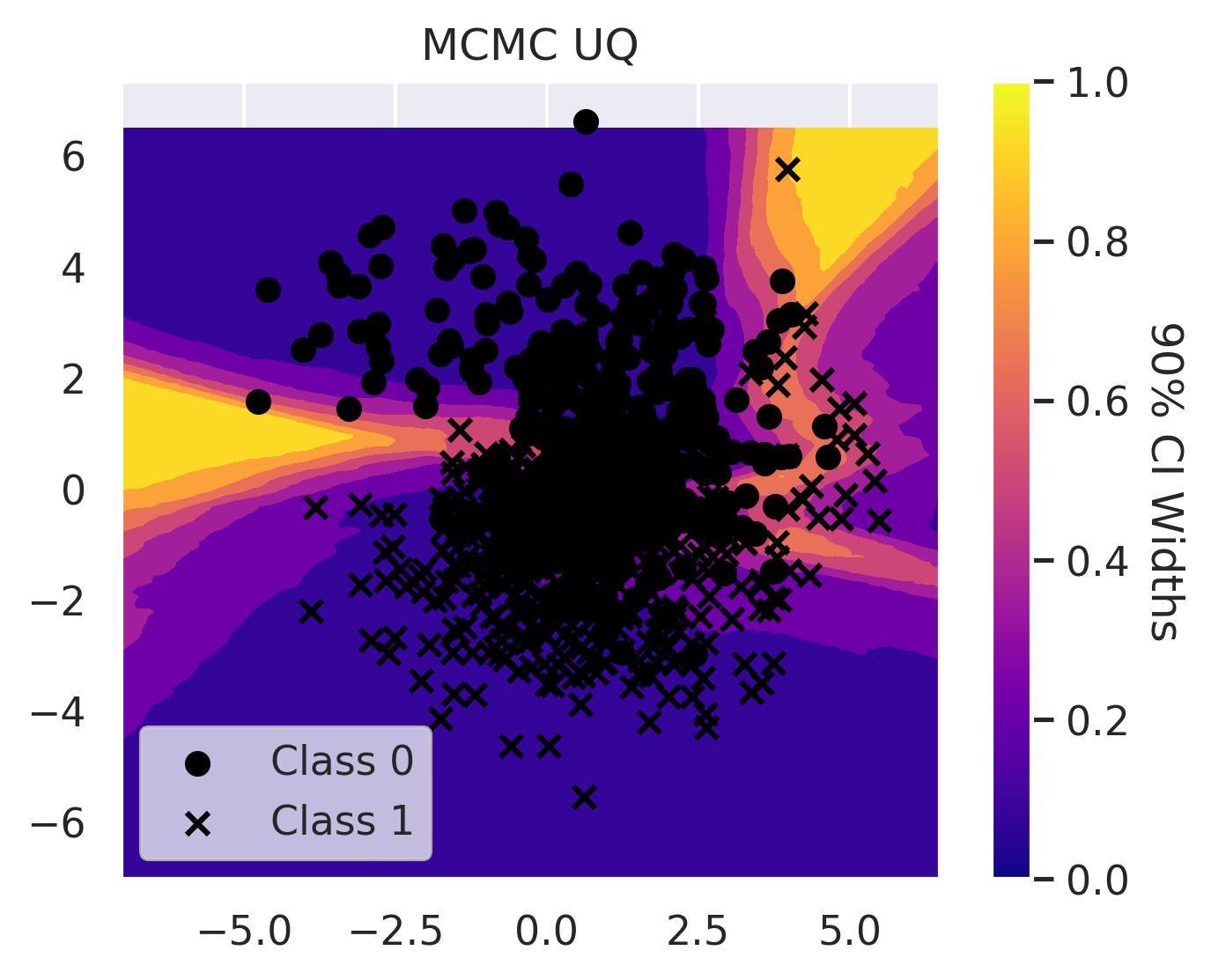} \hfill 
\includegraphics[width=0.3\textwidth]{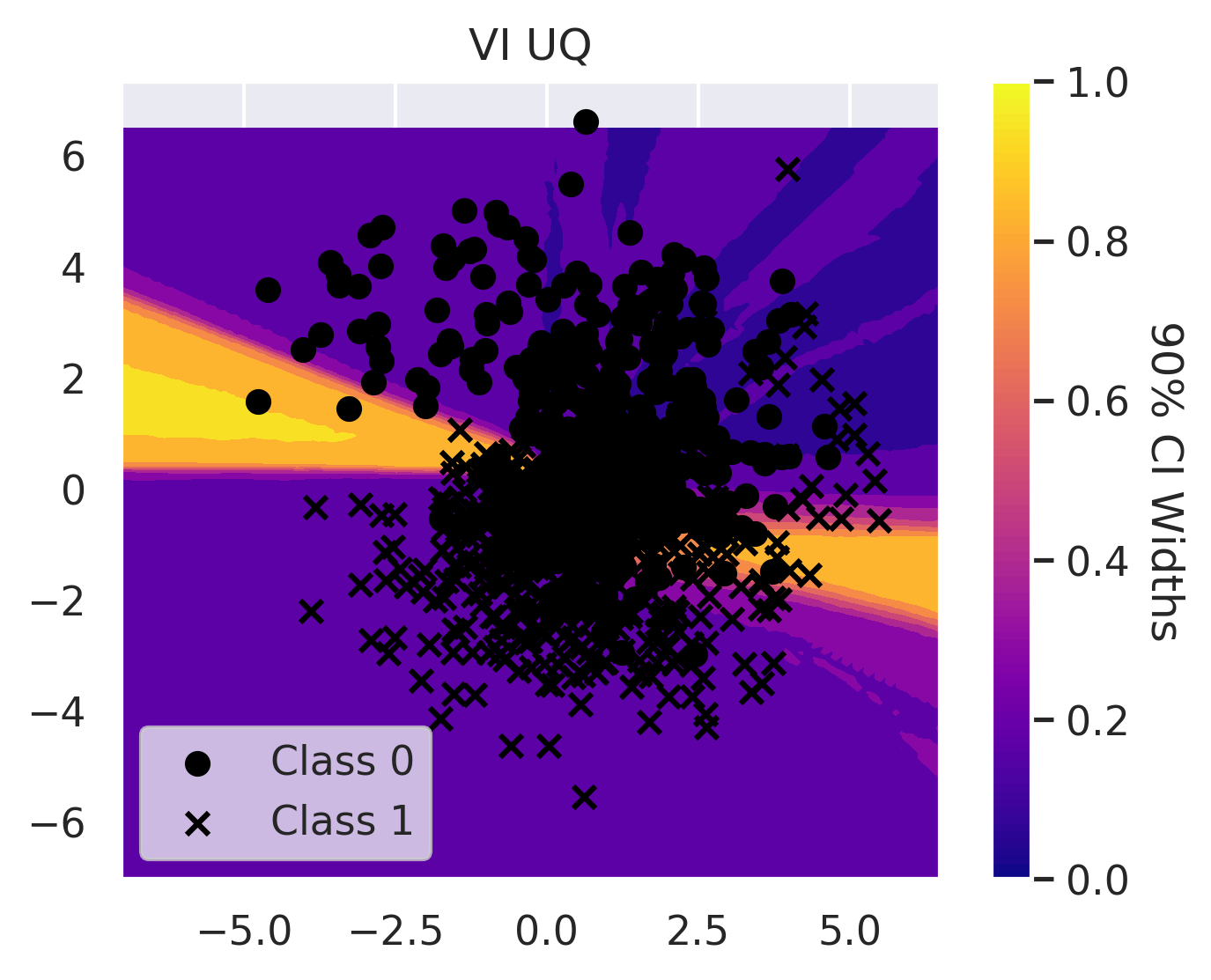} \hfill 
\includegraphics[width=0.3\textwidth]{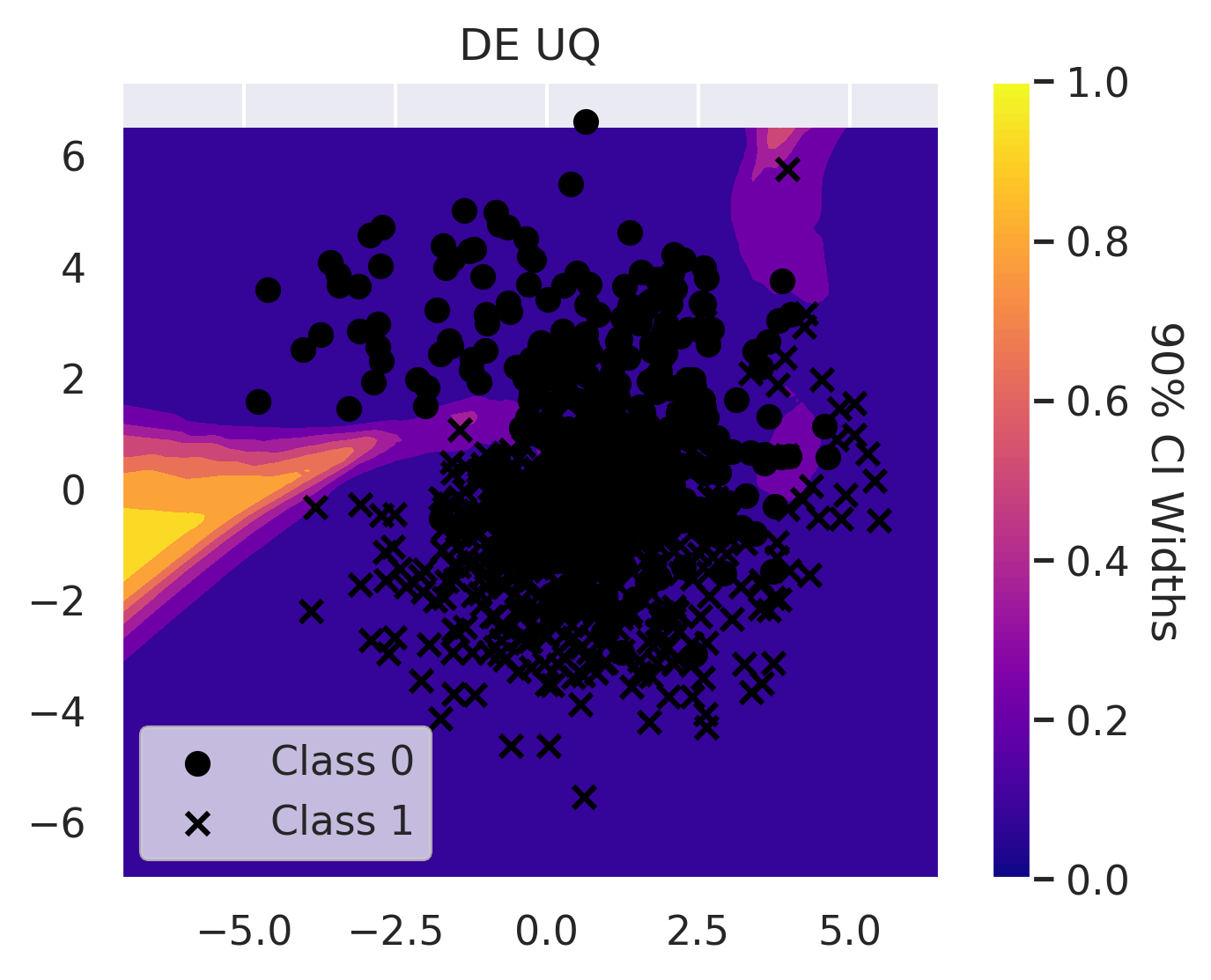} \hfill 
\includegraphics[width=0.3\textwidth]{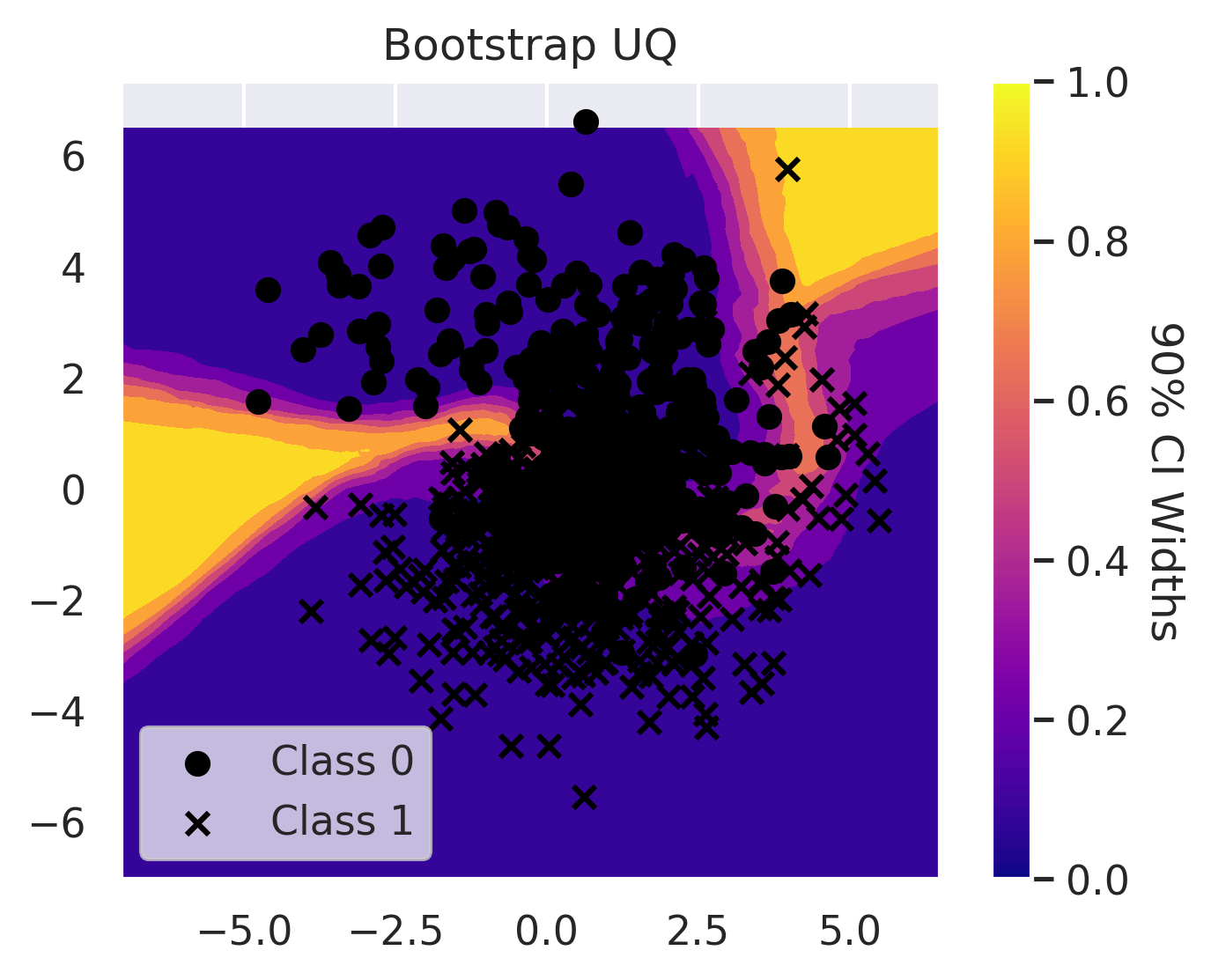} \hfill 
\includegraphics[width=0.3\textwidth]{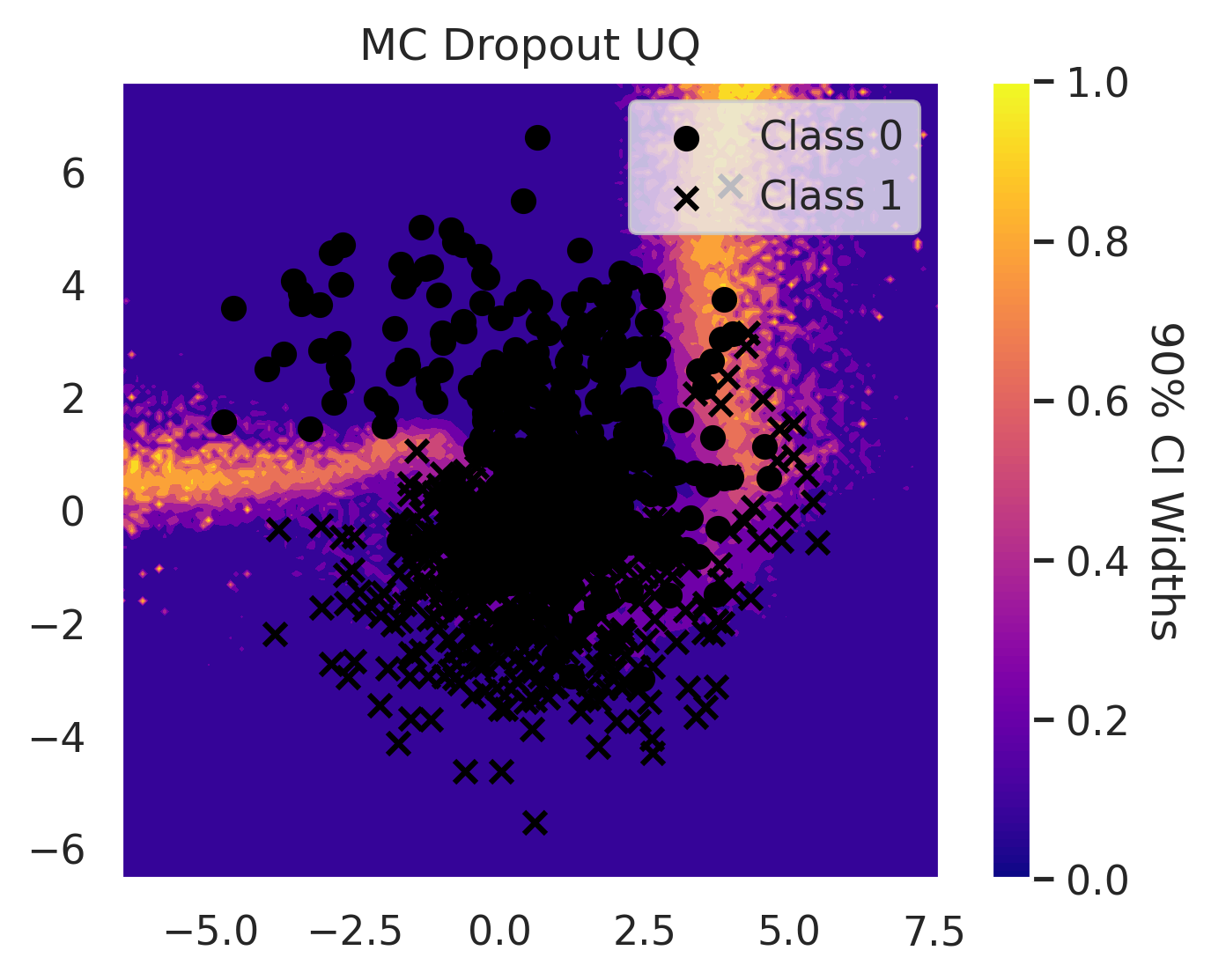} \hfill 
\includegraphics[width=0.3\textwidth]{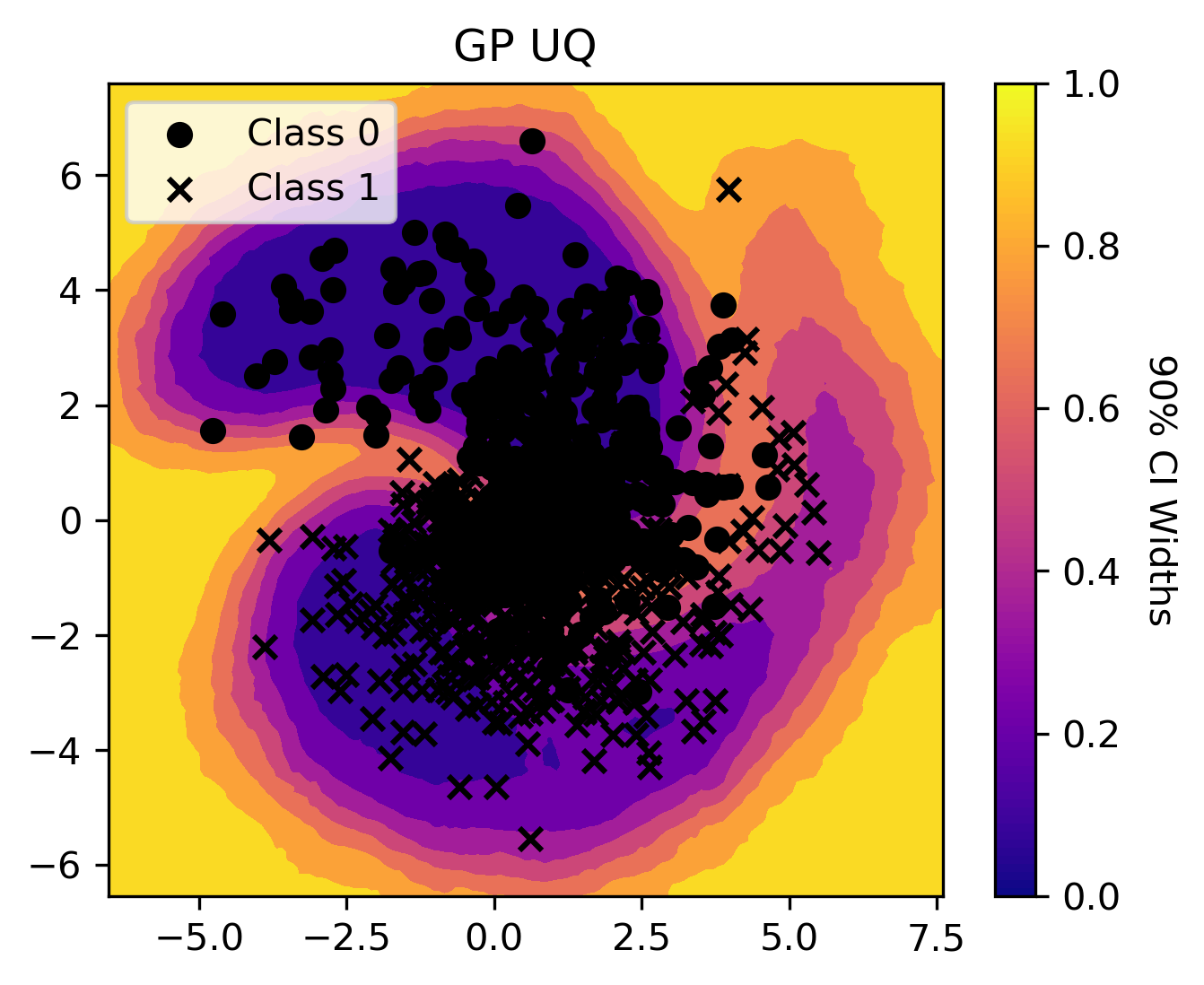} \hfill 
\caption{Uncertainties for each model via 90\% credible interval widths on one TCC simulation. Training data is overlaid.}
\label{fig:TC_mcmc_uq}
\end{figure*}

Figure \ref{fig:TC_mcmc_pred} shows the prediction surface for one simulated TCC data set for each model. Figure \ref{fig:TC_mcmc_uq} shows the width of a 90\% credible interval for one simulated TCC data set for each model. The estimation surfaces for all methods except the GP are similar. The GP appears to also be measuring the density of the domain as well as class probabilities, potentially giving it an OOD measure. The interval widths among all the methods except GP are also similar. The main difference of the DL models is that the DE and MC dropout uncertainty doesn't fan out as quickly as it departs from training data. This behavior is expected since DE does not account for sampling variation. The MCMC and bootstrap plots look similar, and based on the metrics in Table \ref{tab:simresults}, they are the most reliable NN models. 

\section{Discussion}

There are several results from the simulations that are worth further discussion. First, DE failed to provide an accurate measure of the full uncertainty in the simulation. Although the model was well calibrated (as measured by ECE) compared to other models, its credible intervals undercovered the nominal rate indicating it is not measuring epistemic uncertainty correctly. This is not surprising since DE creates an ensemble by simply using different starting values for each model in the ensemble. Practically this means the uncertainty the ensemble is capturing is the optimization uncertainty. Although this may be of interest in some scenarios, we do not believe this is the case for most users. However, DE is a simple way to understand the complexity of the training procedure. In \cite{laka2017}, the authors say for that there is little difference between DE and bootstrap when training sets are large. However, in cases where we are not data-rich, as in many high-consequence national security problems, we do not have the luxury of an abundance of data. Therefore, for high-consequence problems, we recommend to proceed with caution when using DE, and urge users to understand theoretically which types of uncertainty DE will measure, and which it will not.

Simply resampling data with replacement (bootstrap) for each model in the ensemble gives a theoretically plausible solution to the simplicity of DE. The bootstrap performed only slightly worse than BNN-MCMC, giving reasonable coverage with relatively skinny interval widths and comparable ECE. This additional step requires no additional computational burden compared to DE.

Bayesian neural networks fit using MCMC significantly outperformed BNN fit using VI. Although MCMC is the gold standard for Bayesian estimation, we hoped VI would have given better results given the theoretical guarantees it has. We do note that BNN fit with VI is still a difficult process, and we believe it is possible better results could be obtained using different software or VI algorithms. But in light of this, we  recommend caution for non-experts using BNN fit via VI. VI provides a significant speedup that should not be ignored, therefore future work should continue to develop VI algorithms and continue to make them more user-friendly. More research and applications of BNN fit using VI will help understanding of how to diagnose common training issues. 

We can now tie these low dimensional, oracle-like evaluations of UQ quality back to our original high consequence application in Section \ref{sec:application}. Although not a causal relationship, poor results in low-dimentional simplistic examples are often an indicator that algorithms will not improve as the data and models become more complex. The originally proposed HCSs are predicated upon the notion of good quality UQ estimates, yet our simulation results indicate low-quality UQ estimates from DE and VI (and MC dropout to a lesser degree). This would suggest that for more complex modeling tasks, VI and DE UQ estimates are likely to be of lower quality than a model such as the MCMC BNN or bootstrap NN.

There are ample opportunities for future work in the assessment of the quality of UQ for DL models. New metrics should be created that assess the quality of UQ given by DL models, preferably ones that are more well suited to the DL framework. Although the traditional statistical metrics used in this paper are adequate, there are certainly better approaches. We also argue for metrics  beyond  combining the two, such as with the coverage width criterion of \cite{khosravi2011} or evaluating coverages at a large number of nominal rates such as with the continuous ranked probability score from \cite{zamo2018}. We recognize these metrics are useful in evaluation too, but they still require knowing the underlying \emph{true} probability distribution, which for classification problems is only possible with simulated data. New metrics will be able to be used on real data to compare which UQ method to use for that specific data set, much like model selection is currently done (where selection only considers predictive performance of the model).   A metric analogue to the AIC, which allows simple comparison of model fits, is desired to measure the quality of UQ.


\section{Conclusion}

Uncertainty quantification of DL models is an active area of research since researchers and users of DL models have realized point predictions are not always enough, especially in high consequence problems. Many different approaches to UQ for DL models have been proposed, however, there has been little research into the \emph{quality} of those UQ methods. We fit several UQ for DL models on a target detection application, but looking only at the predictions and uncertainties from the models does not tell a decision maker which model best captures the underlying uncertainty. In fact, it introduces more questions than answers.  
In an attempt to answer these questions, this paper explores the quality of UQ given by several probabilistic UQ models, including BNN, GP, DE, MC dropout, and bootstrapped NN, using traditional statistical metrics of frequentist coverage and CI width, as well as ECE. A two class classification data set, for which complete knowledge of the data generating mechanism was known, was used to quantitatively assess the UQ qualities. 

BNN trained via MCMC was the clear winner, but this comes with a heavy computational cost. The bootstrap came in a close second and may be more practical to use. It requires the same computation as the popular DE, but appears to provide higher quality UQ. However, this paper only explores two specific cases and therefore more research in this area is needed, and better UQ metrics need to be developed to definitively compare UQ in DL methods.


\section*{Acknowledgements}
This work was supported by the Laboratory Directed Research and Development program at Sandia National Laboratories, a multimission laboratory managed and operated by National Technology and Engineering Solutions of Sandia LLC, a wholly owned subsidiary of Honeywell International Inc. for the U.S. Department of Energy’s National Nuclear Security Administration under contract DE-NA0003525. This paper describes objective technical results and analysis. Any subjective views or opinions that might be expressed in the paper do not necessarily represent the views of the U.S. Department of Energy or the United States Government. SAND2022-8993 C.
The authors would like to thank Michael Darling and Lekha Patel for their contributions in reviewing and improving this paper.

\bibliography{bibliography}


\end{document}